\documentclass[letterpaper, 10 pt, conference]{ieeeconf}  

\IEEEoverridecommandlockouts                              %

\usepackage{amsmath,amsfonts}
\usepackage{algorithmic}
\usepackage{algorithm}
\usepackage{array}
\usepackage[caption=false,font=normalsize,labelfont=sf,textfont=sf]{subfig}
\usepackage{textcomp}
\usepackage{caption}
\usepackage{stfloats}
\usepackage{url}
\usepackage{verbatim}
\usepackage{graphicx}
\usepackage[normalem]{ulem}
\usepackage{tabularray}
\usepackage{CJKutf8}
\usepackage{color}
\usepackage{cite}
\usepackage{hyperref}
\usepackage{graphicx}
\usepackage{multirow}

\begin{document}
\begin{CJK}{UTF8}{gbsn}

\title{\textbf{MuxHand: A Cable-driven Dexterous Robotic Hand Using Time-division Multiplexing Motors}}

\author{Jianle Xu{$^{1*}$}, Shoujie Li{$^{1*}$}, Hong Luo{$^1$}, Houde Liu{$^{1\dagger}$}, Xueqian Wang{$^1$}, Wenbo Ding{$^1$}, Chongkun Xia{$^2$}
\thanks{{$*$} These authors contributed equally to this
work.}
\thanks{This work was supported by the National Natural Science Foundation of China (92248304), Shenzhen Science Fund for Distinguished Young Scholars (RCJC20210706091946001), Shenzhen Science and Technology Program (JCYJ20220530143013030) and Shenzhen Higher Education Stable Support Program  (WDZC20231129093657002).}
\thanks{{$1$} Jianle Xu, Shoujie Li, Hong Luo, Wenbo Ding,  Houde Liu, and Xueqian Wang are with Shenzhen International Graduate School, Tsinghua University, Shenzhen 518055, China.}
\thanks{{$2$} Chongkun Xia is with School of Advanced Manufacturing, Sun Yat-sen University, shenzhen 518107, China}
\thanks{{$\dagger$} Corresponding authors: Houde Liu (liu.hd@sz.tsinghua.edu.cn).}
\thanks{This paper has supplementary material available at \href{https://xujianle.github.io/MuxHand.github.io}{https://xujianle.github.io/MuxHand.github.io}.}}

\markboth{Journal of \LaTeX\ Class Files,~Vol.~14, No.~8, August~2021}%
{Shell \MakeLowercase{\textit{et al.}}: A Sample Article Using IEEEtran.cls for IEEE Journals}

\maketitle

\begin{abstract}

The robotic dexterous hand is responsible for both grasping and dexterous manipulation. The number of motors directly influences both the dexterity and the cost of such systems. In this paper, we present MuxHand, a robotic hand that employs a time-division multiplexing motor (TDMM) mechanism. This system allows 9 cables to be independently controlled by just 4 motors, significantly reducing cost while maintaining high dexterity. To enhance stability and smoothness during grasping and manipulation tasks, we have integrated magnetic joints into the three 3D-printed fingers. These joints offer superior impact resistance and self-resetting capabilities.
We conduct a series of experiments to evaluate the grasping and manipulation performance of MuxHand. The results demonstrate that the TDMM mechanism can precisely control each cable connected to the finger joints, enabling robust grasping and dexterous manipulation. Furthermore, the fingertip load capacity reached 1.0 kg, and the magnetic joints effectively absorbed impact and corrected misalignments without damage.

\end{abstract}

\section{Introduction}

The robotic hand, as a general end-effector for humanoid robots, performs both grasping\cite{li2022tata, li2021design} and dexterous manipulation tasks\cite{armada2005introductory}. Numerous advanced dexterous hands have been developed, such as the DLR/HIT Hand \cite{HITDLR}, the ILDA Hand \cite{ILDAhand}, the RoboRay Hand \cite{roboray}, and others\cite{robonaut, UtahMIT, shadow_dexterous_hand}. These hands generally have a high degree of freedom (DOF), however, an increased DOF often requires more motors, which results in a larger size and higher cost.
\begin{figure}
	\centering
  \includegraphics[width=0.48\textwidth]{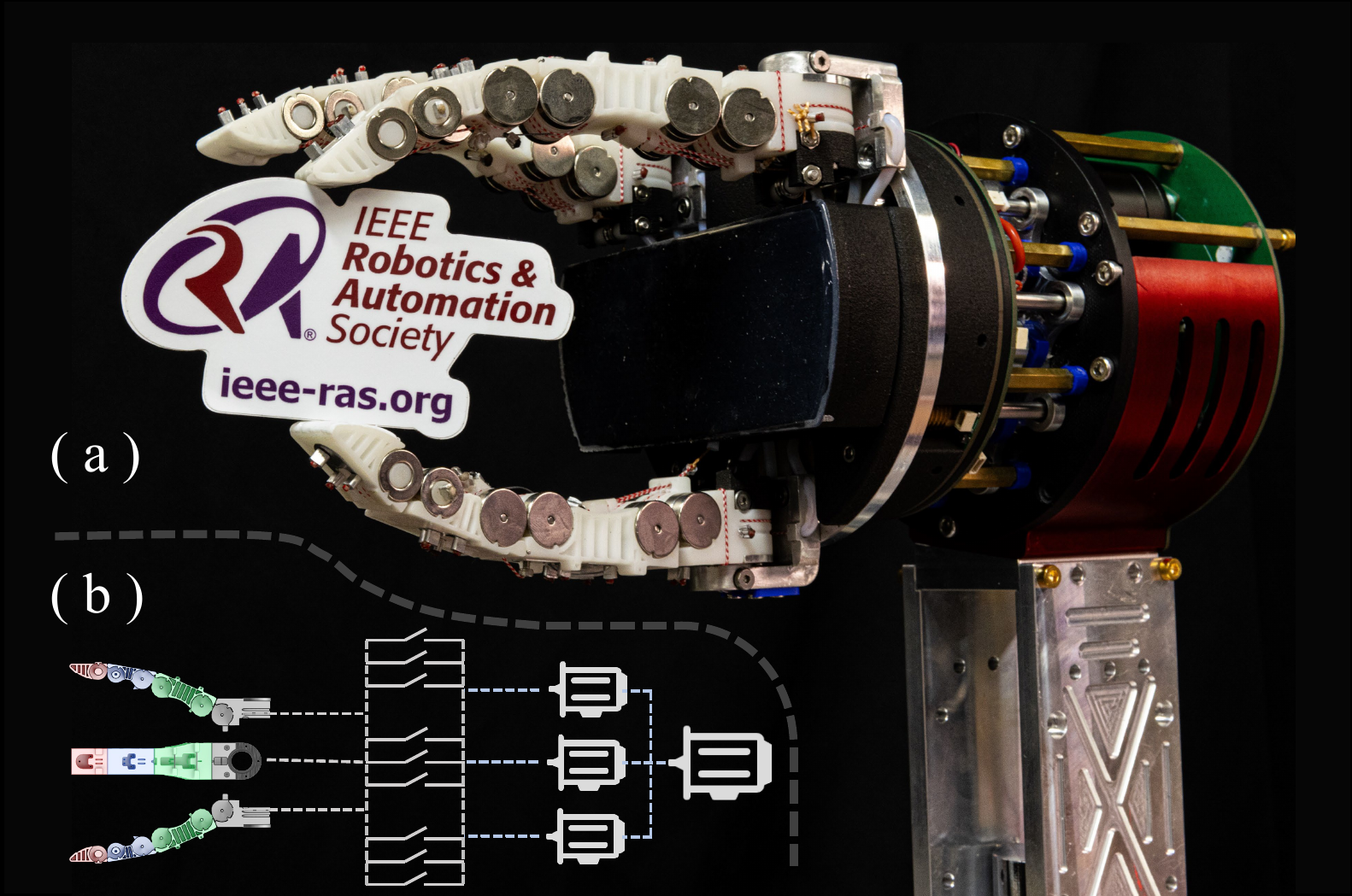}

	\caption{ MuxHand, a cable-driven dexterous robotic hand using time-division multiplexing motors. (a) MuxHand prototype. (b) Simplified system schematic. } \label{fig:1}
 \vspace{-0.5cm}
\end{figure}

An alternative approach involves the use of underactuated mechanisms. The concept of utilizing fewer motors to achieve greater dexterity in robotic hands has gained significant attention in recent years \cite{sun2021design}. Due to the reduced number of motors, these systems are generally more compact. Furthermore, underactuated mechanisms enable robust grasping of objects with unknown or irregular shapes \cite{sun2021design,yan2022c,della2018toward,gao2023new}. However, a major limitation of such systems is that, with a reduced number of DOF, individual joint control is not possible, making dexterous manipulation challenging.
Therefore, the dexterity of robotic hands is often proportional to the number of motors. Ideally, the number of motors should match the DOF of the hand \cite{kim2019fluid},\cite{zhu2022anthropomorphic}. However, motors are usually the most expensive and space-consuming components, making it a significant challenge to drive a robotic hand with a minimal number of motors while maintaining sufficient dexterity.

While underactuated mechanisms reduce the size and cost of dexterous hands, independently controlling each joint remains essential. Kontoudis \textit{et al}. \cite{kontoudis2015open} and Baril \textit{et al}. \cite{baril2013design} introduce mechanisms that utilizes mechanical selectors to connect or disconnect a motor from a specific cable, but they rely on manual button activation, limiting their applicability. Kim \textit{et al}. proposed a switchable cable-driven (SCD) mechanism that uses multiple electrostatic clutches to individually control several cables with a single motor. However, the friction force of these electrostatic clutches is insufficient to hold cables under high tension, and the system required significant space.

To overcome these limitations, we developed a novel mechanism for switching cable-driven actuation. This design uses a small number of motors to dynamically switch output positions at different times, enabling individual control of specific cables while minimizing size and maintaining dexterity.
In this paper, we present MuxHand, a robotic hand that utilizes the time-division multiplexing motor (TDMM) mechanism to control the 9 DOF of its three fingers with just 4 motors. MuxHand is composed of a compact drive box and a cable-driven finger module.
The compact drive box, designed based on the TDMM mechanism, integrates 4 motors, a gear drive module, an electromagnetic coil module, a worm and worm wheel module, and six printed circuit boards (PCB). These components are highly integrated within the drive box to maintain a minimal footprint.
The cable-driven finger module comprises three fingers, each 3D printed using photosensitive resin\cite{ma2013research}. We embedded magnets with specialized magnetization directions in each finger joint, providing the fingers with exceptional impact resistance and self-resetting capabilities.

The contributions of our work are as follows: 

\begin{itemize} 
\item[$\bullet$] We design a cable-driven robotic hand using a TDMM mechanism, which controls 9 DOF with just 4 motors, reducing cost and size. 
\item[$\bullet$] We design a cable-driven finger with magnetic joints that provide impact resistance and self-resetting capabilities, enabling robust grasping and dexterous manipulation.
\item[$\bullet$] We validate MuxHand's performance through experiments in grasping, dexterous manipulation, and fingertip load tests, demonstrating the effectiveness of the TDMM mechanism and magnetic joints. 
\end{itemize}

\section{Mechanical Design}

In this section, we will describe the structure of MuxHand and provide a detailed explanation of the TDMM mechanism. MuxHand consists of two primary components: a compact drive box (Fig. 2), which utilizes the TDMM mechanism, and a cable-driven finger module (Fig. 5), featuring a fully decoupled joint structure.

\begin{figure*}
	\centering
  \includegraphics[width= 0.96\textwidth]{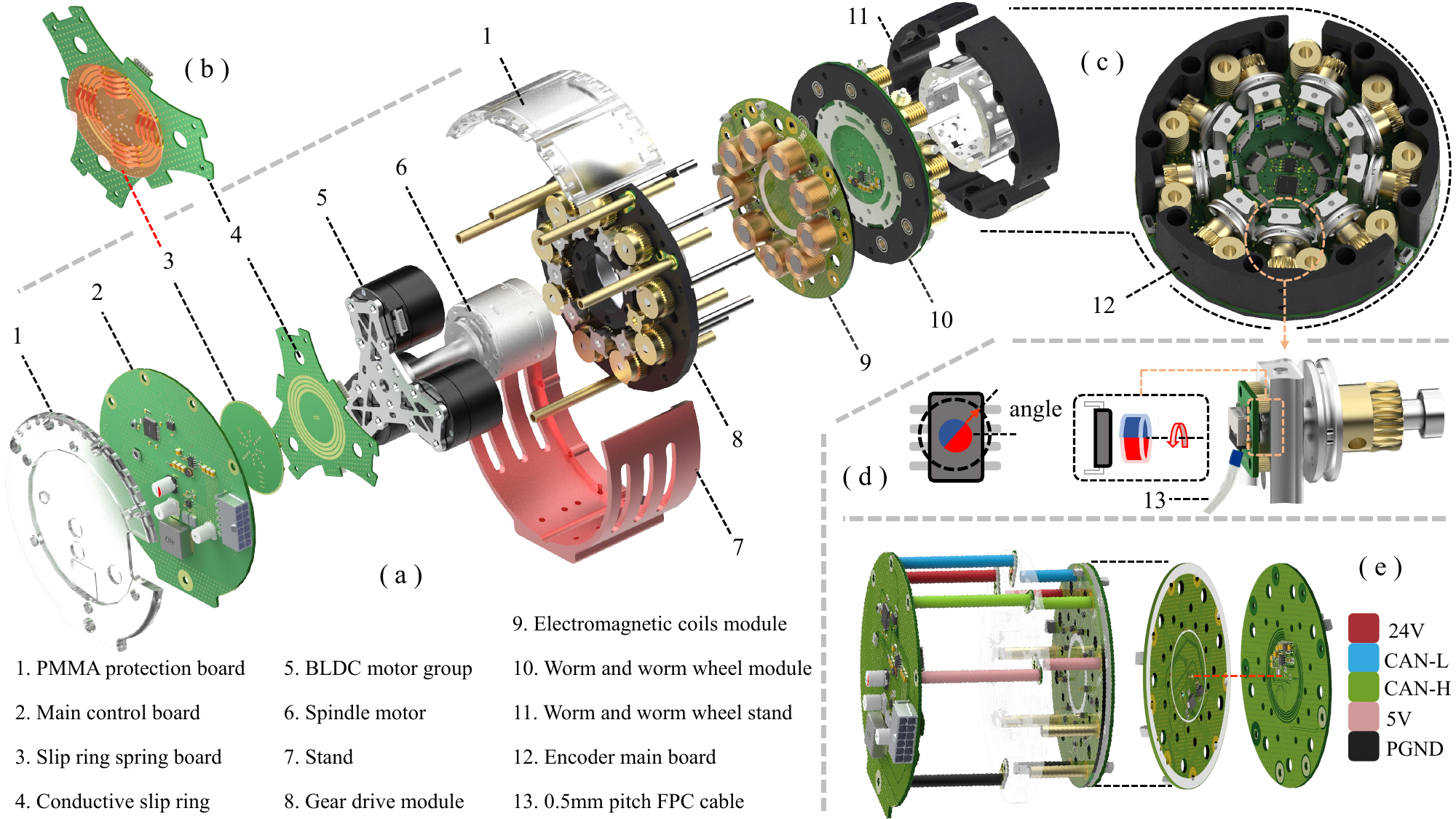}
	\caption{Drive box structure. (a) Exploeded view of the drive box. (b) Slip ring structure. (c) Structure of the magnetic encoder groups. (d) Schematic and structure of the encoder. (e) Pathway for power and data transmission. } \label{fig:2}
 \vspace{-0.5cm}
\end{figure*}

\subsection{Drive Box Design}

To effectively control MuxHand, we develope a drive box based on the TDMM mechanism, utilizing four Brushless Direct Current (BLDC) motors to individually drive nine cables connected to the finger joints. The drive box measures 200 mm in height and 120 mm in diameter. To achieve a more compact structure, six highly integrated PCB are designed, as shown in Fig. 2(a). These include the main control board (MCB), slip ring spring board (SRSB), conductive slip ring (CSR), electromagnetic coil module board (ECMB), encoder main board (EMB), and encoder board (ECB).

The MCB manages the entire hardware system, handling communication with other boards and sensors, as well as task planning and power conversion. It supports up to 60V direct current input and uses the STM32F446-RET6 Microcontroller Unit (MCU) with a 180MHz  frequency.
To prevent the wires supplying power and communication signals to the BLDC motor module from tangling due to spindle rotation, a conductive slip ring structure is designed (Fig. 2(b)). This ensures uninterrupted power and communication to the BLDC motor module.

To implement the TDMM mechanism, we develope a system to control the shaft torque output at different times. This system includes an electromagnetic coil module and a shaft module equipped with magnets. The shaft module consists of 9 individual shafts, with a single shaft structure shown in Fig. 4(b).

To accurately measure the angle of each finger joint, we design a compact magnetic encoder structure. As shown in Fig. 2(d), the schematic illustrates the magnetic encoder. We use the AS5047P magnetic encoder chip, which offers a 14-bit resolution with a range from 0 to 16,383. To capture the joint angles, a radially magnetized magnet is mounted on the shafts of the worm wheel, and an encoder control board (ECB) is designed to integrate these components. Since MuxHand has three fingers with a total of 9 DOF, nine sets of this magnetic encoder structure are required. The structure of the magnetic encoder arrays is shown in Fig. 2(c).
All encoder boards are connected to the encoder main board (EMB) in Fig. 2(a) using 0.5 mm pitch FPC cables\cite{koh2016radiated}. Each ECB communicates with the EMB via the Serial Peripheral Interface (SPI) protocol at a baud rate of 9.0 Mbits/s. Once the EMB collects the angle data, its MCU transmits the data to the CAN bus\cite{nahas2009reducing}, which operates at a baud rate of 1 Mbits/s.

Additionally, we consider power supply and data transmission challenges due to the limited space inside MuxHand. Arranging separate power and communication cables for each PCB can lead to tangled wires during operation, causing instability. As illustrated in Fig. 2(e), the power supply and data transmission path is designed to address this issue. To simplify the design, we utilize ten copper columns as pathways for power and data, while also serving as supporting structures. The colored blocks in Fig. 2(e) represent the function of each copper column. The red, blue, green, pink, and black columns represent the 24V power supply, CAN-L transmission path, CAN-H transmission path, 5V power supply, and power ground (PGND), respectively.
Due to space constraints, a DC-DC converter for supplying 3.3V to the MCU could not be embedded in the electromagnetic coil module. To resolve this, we use a pogo pin\cite{ehtiatkar2011mechanical} (A small spring-loaded connector commonly found in electronic devices) to transfer 3.3V power between the electromagnetic coil module and the EMB. The red line in Fig. 2(e) indicates the corresponding position of the pogo pin.

\begin{figure*}
	\centering
  \includegraphics[width=1.0\textwidth]{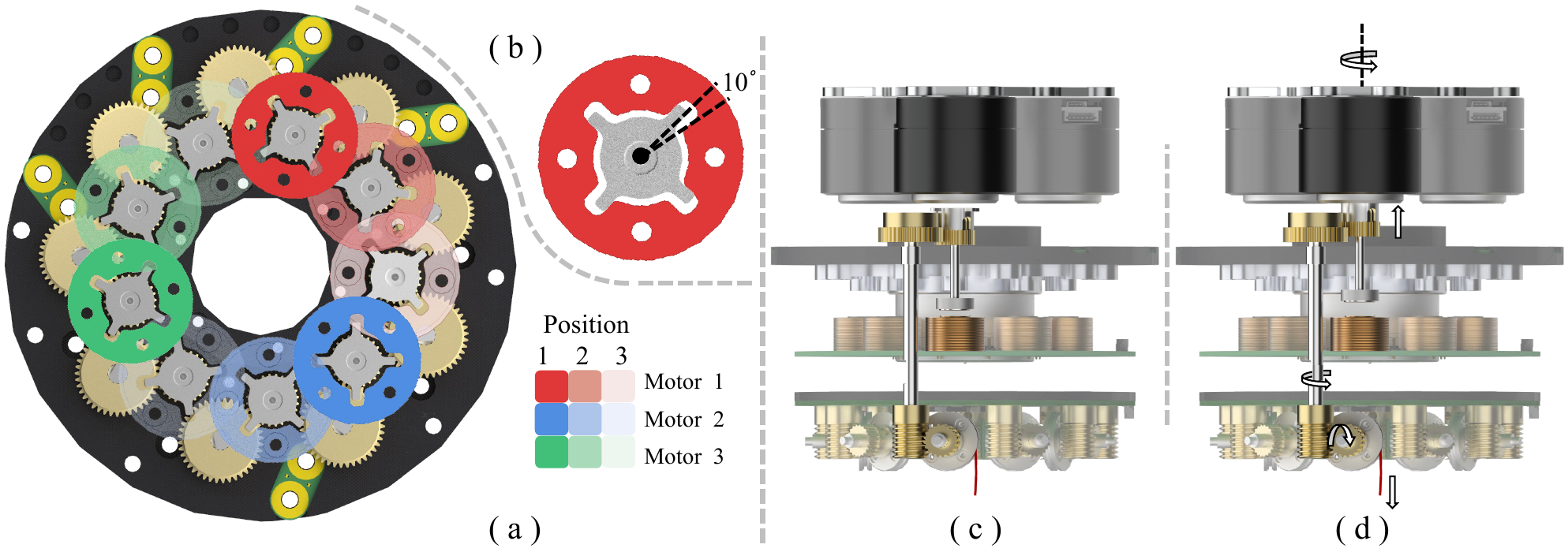}
	\caption{TDMM mechanism and transmission mechanism. (a) Operating positions of the BLDC motor group at different times. Different colors represent individual motors, with varying shades of the same color indicating the motor positions at different time points. (b) Maximum allowable angular error between the slot and plug. (c) BLDC motor group reaching the target position. (d) Torque transmission path of the BLDC motor.} \label{fig:3}
 \vspace{-0.5cm}
\end{figure*}

\subsection{TDMM Mechanism And Transmission Mechanism}

The principle of the TDMM mechanism we propose is similar to that of gear shifting. It consists of a BLDC motor group, a spindle motor (HTM-4538-20-NE), a shaft module, and an electromagnetic coil module. The spindle motor rotates the BLDC motor group to different positions, as shown in Fig. 3(a), which illustrates the BLDC motor group’s state. The BLDC motor group includes Motor 1, Motor 2, and Motor 3. In Fig. 3(a), different colors represent different motors, and varying shades of the same color indicate the positions of each motor at different times. 

Initially, when no current flows through the electromagnetic coil, the iron core is not magnetized but still attracts the plug, which has a magnet attached to it. The left part of Fig. 4(b) illustrates this state. Once the BLDC motor group reaches the target position, as shown in Fig. 3(c), the electromagnetic coil at this position is energized in a fixed direction, causing the iron core to become magnetized. According to the right-hand rule, the magnetic polarity of the iron core is determined. Since the plug has an attached magnet, it is lifted by the repulsion force from the magnetized iron core, aligning the gear component with the motor output slot, the right part of Fig. 4(b) shown this process. Consequently, as shown in Fig. 3(d), which depicts the torque transmission path of a single motor, the torque output by the BLDC motor group is transmitted through the gear drive module and the worm and worm wheel module, ultimately reaching the winding wheel.

During this process, although we use closed-loop speed and position control to drive the spindle motor to different positions, there is inevitably some mechanical error between the torque transmission slots and plugs. As shown in Fig. 3(b), after the plug is lifted into the slot, the maximum error between the slot and plug is approximately 10°. However, this 10° error occurs at the driving motor, and when translated to the winding wheel, it must be adjusted through the reduction gears.
Fig. 4(a) presents a simplified schematic of the transmission mechanism, where the arrows indicate the direction of gear movements. Based on this schematic, we can calculate the transmission ratio $k$:
\begin{eqnarray}
k = \frac{Z_2 \cdot Z_4}{Z_1 \cdot Z_3 } .
\end{eqnarray}

where $Z_1,Z_2,Z_3$ and $Z_4$ represent the number of teeth on each gear, with the gear colors corresponding to the color coding for the number of teeth. The final calculated reduction ratio is 33.33, which results in a maximum error of only 0.30° when apply to the winding wheel.

\begin{figure}
	\centering
  \includegraphics[width=0.45\textwidth]{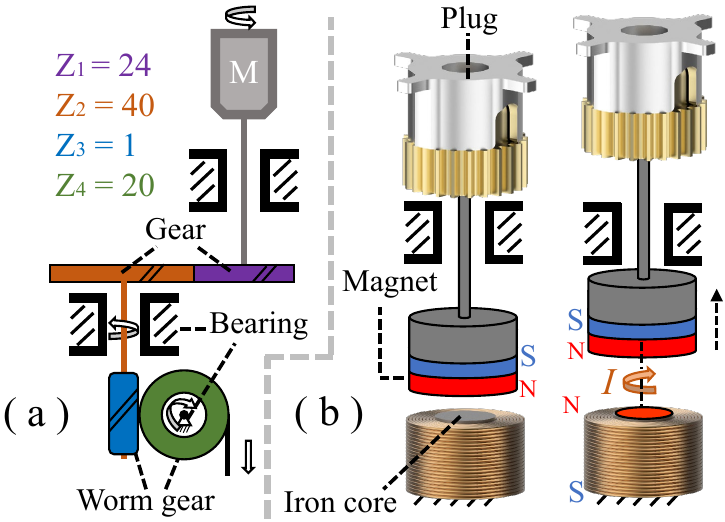}
	\caption{Simplified schematic of gear transmission and plug movement. (a) Gear transmission. The color of the gears corresponds to the color coding of the number of gear teeth. (b) Schematic of plug movement. S and N represent the south and north poles of the magnet, respectively, and $I$ represents the direction of the current. } \label{fig:4}
  \vspace{-0.5cm}

\end{figure}

\subsection{Finger Module Design}
 
To enhance the dexterity of MuxHand, we design fingers with fully decoupled joints, ensuring that the movement of one joint does not affect the others\cite{lee2008design}. MuxHand comprises three identical fingers, each with 4 degrees of freedom (DOF), as shown in Fig. 5(f). Of these DOF, three are actively controlled, while the remaining one is coupled with the Proximal Interphalangeal (PIP) joints. The entire finger is 3D printed using photosensitive resin material, and each finger's joints are cable-driven. To reduce weight while maintaining strength, the fingers feature a hollowed-out structure.
In a cable-driven system, minimizing friction between the cable and its constraints is crucial for improving transmission efficiency and output force\cite{tan2024control}. To address this, we installed polytetrafluoroethylene (PTFE) tubing at key positions, which has a very low coefficient of friction\cite{RADULOVIC20141133}.

The finger consists of three types of cables: coupling cables (Fig. 5(a)), constraint cables (Fig. 5(b)), and drive cables (Fig. 5(c)). Fig. 5(d) illustrates the positioning of all the cables within the finger. The coupling cables allow the Distal Interphalangeal (DIP) joint to rotate synchronously with the PIP joint at a specific ratio, mimicking the movement of a human finger. Constraint cables are used to restrict and connect the various finger joints, while drive cables are used to actuate the joints.
Since all drive cables must pass through the MetaCarpoPhalangeal (MCP) roll joint, it is crucial to ensure that there is no interference between the cables during finger joint movements and no coupling effects between the drive joints. Therefore, the cable routing needs to be decoupled. Fig. 6(e) shows the decoupled routing path of the drive cables: the orange cable represents the PIP joint drive cable, the yellow cable represents the MCP pitch joint drive cable, and the green cable represents the MCP roll joint drive cable.

\begin{figure}
	\centering
  \includegraphics[width=0.45\textwidth]{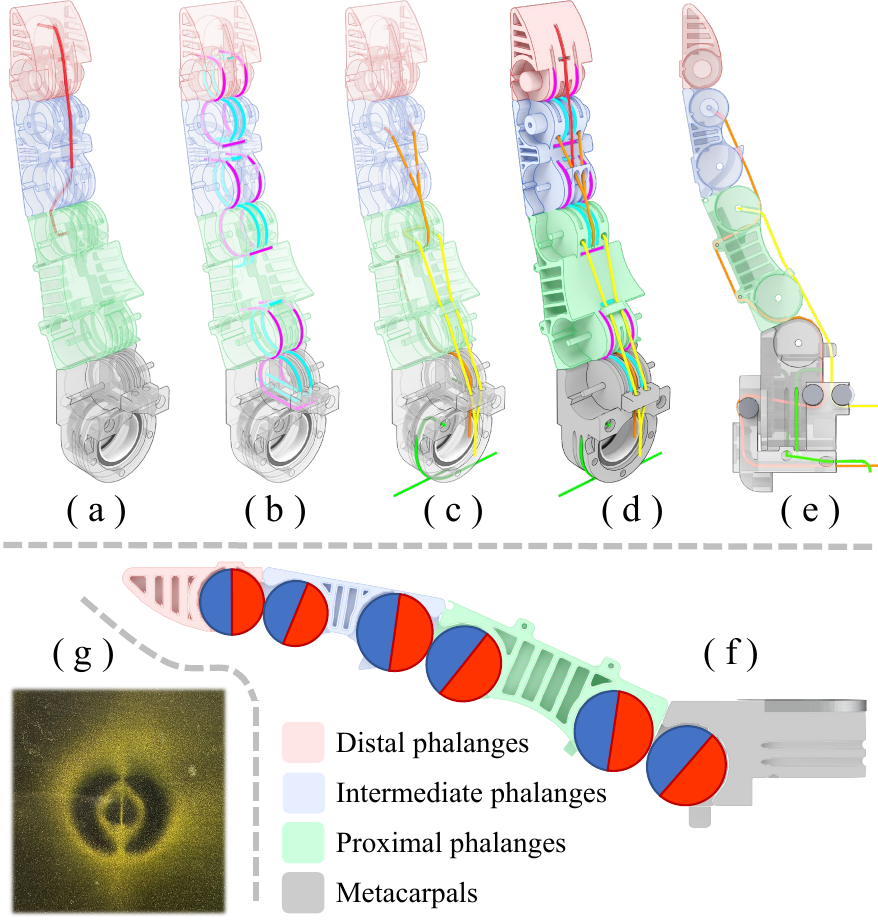}
	\caption{Structure of the finger. (a) DIP joint and PIP joint coupling cable. (b) Constraint cables between the joints of each finger. (c) The orange cable represents the PIP joint drive, yellow is for the MCP pitch joint drive, and green is for the MCP roll joint drive. (d) Overall cable routing. (e) Decoupling principle of the drive cable at the MCP joint. (f) Magnet arrangement in each finger joint. (g) Magnetic pole distribution is displayed using a magnetic field viewing card.} 
 \label{fig:5}
   \vspace{-0.5cm}
\end{figure}

Fig. 6(b) illustrates the schematic of the coupling relationship between the DIP joint and PIP joint, where the $r_1$ and $r_2$ are the radii of the rolling joints of the DIP joint and PIP joint, respectively, and the $r_3$ is the radius of the winding wheel. The yellow line in Fig. 6(b) is the drive cable for the PIP joint. The change in length of this cable caused by the winding wheel is denoted as $\Delta x_2$. When the winding wheel rotates by an angle $\varphi_1$, $\Delta x_2$ is given by:
\begin{eqnarray}
    \Delta x_2 = \varphi \cdot r_3  .
\end{eqnarray}
Since the PIP drive cable is fixed in the intermediate phalanges and is constrained to be tangent to the rolling joint surface of the proximal phalanges, the rotation angle of the PIP joint $\theta_2$ can be derived from geometric relationships as:
\begin{eqnarray}
    \theta_2 = 2\alpha_2  .
\end{eqnarray}
Here, $l_2$ can be determined as:
\begin{eqnarray}
    l_2 = \sqrt{r_2^2 + (2r_2 - \Delta x_2)^2} ,
\end{eqnarray}
where $\alpha_2$ is the angle of rotation of the virtual link of the proximal phalanges relative to the PIP joint, given by:
\begin{eqnarray}
    \alpha_2 = \arcsin(\frac{5r_2^2 - l_2^2}{4r_2^2}) .
\end{eqnarray}
Substituting $l_2$ into this with (4),
\begin{eqnarray}
    \theta_2 = 2\alpha_2 = 2\arcsin[\frac{\Delta x_2}{r_2} - (\frac{\Delta x_2}{2r_2})^2],
\end{eqnarray}
with  $\theta_2$,  the change in length of the coupling cable $\Delta x_1$, as shown in Fig. 7(a), can be calculated. This change determines the rotation angle of the DIP joint:
\begin{eqnarray}
    \Delta x_1 = 2l_2\sin(\alpha_2).
\end{eqnarray}
Using a similar method, the rotation angle of the DIP joint $\theta_1$ can be calculated as:
\begin{eqnarray}
    \theta_1 = 2\alpha_1 = 2\arcsin[\frac{\Delta x_1}{r_1} - (\frac{\Delta x_1}{2r_1})^2].
\end{eqnarray}
The calculation method for the MCP pitch joint is similar to that for the PIP joint. As illustrated in Fig. 6(c), the figure shows the relationship between the rotation angle of the winding wheel and the DIP, PIP, and MCP pitch joints. And we use linear regression equations to fit the relationship, which shows that there is a certain linear relationship between the rotation angles of the DIP and PIP joints and the winding wheel. This ensures smooth operation of the finger joints.

\begin{figure}
	\centering
  \includegraphics[width=0.48\textwidth]{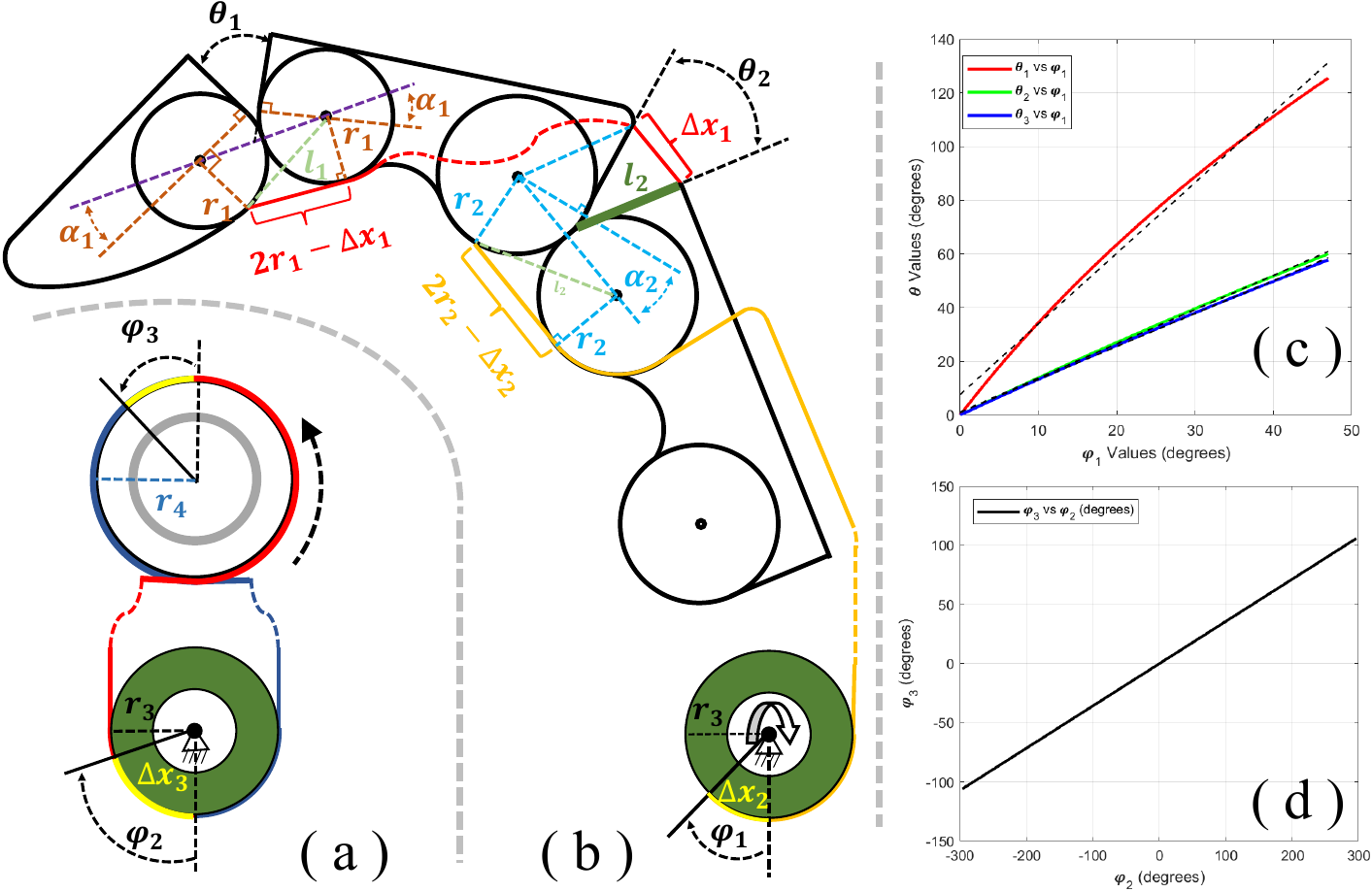}
	\caption{The relationship between the winding wheel and each joint. (a) The relationship between the rotation angle of the winding wheel and the MCP roll joint. (b) The relationship between the rotation angle of the winding wheel and the movement angles of the DIP, PIP, and MCP pitch joints. (c) The relationship between $\theta_1$, $\theta_2$, $\theta_3$, and $\varphi_1$ is determined and fitted using a linear regression equation. Where the $\theta_3$ is the angle of the MCP pitch joint. (d) The relationship between $\varphi_3$ and $\varphi_2$.}
 \label{fig:7}
  \vspace{-0.5cm}
\end{figure}

The calculation method for the MCP roll joint's rotation angle relative to the winding wheel's rotation angle differs slightly from that of the other joints. As shown in Fig. 6(a), when the winding wheel rotates by a certain angle, the MCP roll joint experiences the same arc length displacement, resembling the motion of a pair of gears. Therefore, without accounting for cable tension, this motion follows a completely linear relationship. Fig. 6(d) illustrates the relationship between the winding wheel's rotation and the MCP roll joint.

To provide a reset torque for each joint and ensure stability between connections, we do not to use springs as the reset mechanism. Instead, we employ a specialized radially magnetized magnet in each joint. Fig. 5(g) illustrates the magnetic field distribution of the magnet, which extends radially along the cylinder. Unlike springs, these magnets do not take up additional space and are not prone to fatigue from prolonged or repeated use, significantly enhancing the durability of the finger joints. Furthermore, the radial magnetization allows the magnets to withstand certainly impacts and misalignment without damage, while enabling automatic resetting once external forces are removed. Fig. 5(f) shows the arrangement of the magnets, with red indicating the north pole and blue the south pole.

\section{Experiments and Results}

In this section, we conduct grasping and in-hand manipulation experiments to demonstrate the MuxHand's performance. Additionally, we conduct experiments to verify the fingertip's load capacity and the joints' self-resetting ability when subjected to external forces. Table \uppercase\expandafter{\romannumeral1} presents the overall parameters of the MuxHand.

\begin{table}[htbp]
\centering
\caption{MuxHand Overall Performance Parameters}

\begin{tabular}{c  p{4cm}}
  \hline
    Items  & \multicolumn{1}{c}{Parameters} \\ \hline  
    
    Weight & 2.6 kg \\ 
    
    Dimension (H × L × W) & 264mm × 120mm × 120 mm \\ 
    
    Fingertip load & 1.0 kg \\ 
    
    Type of coupling mechanisms & Gears, worm gears, and cable drive\\ 
    Cable material & 0.6 mm diameter PE cable\\
    Finger material & Photosensitive resin \\
    
    \multirow{2}{*}{Motors} & BLDC motor (0.4 Nm) $\times$ 3  \newline  Spindle motor (1 Nm) $\times$ 1 \\ 
    
   \multirow{2}{*}{Electronics} & STM32F446RET6 $\times$ 1 \newline                  STM32F103CBT6 $\times$ 2 \\

    Communication protocol & CAN (1Mbits/s) \\ 
    
    Maximum Input Voltage & 60 Volt \\
    Maximum Power & 240 Watt  \\ \hline

 \vspace{-0.8cm}

\end{tabular}
\end{table}

\subsection{Grasping Experiments}

The first experiment focus on the MuxHand's ability to grasp objects. We select various objects from the YCB object set\cite{7254318} and common objects in daily life. The YCB object set is a widely recognized benchmark dataset for evaluating robotic grasping, manipulation, and object recognition. Twelve objects are tested, as shown in Fig. 7, including a cracker box, foam brick, USB hub, can of chips, apple, metal mug, thermometer, plastic wine glass, lighter, dice, mustard container, and ice-cream stick, with their respective labels in Fig. 7(a)–(l).

Due to the MuxHand's 3 DoF per finger, it is capable of grasping objects in a variety of ways. The MuxHand can use two fingers to pinch objects or employ three fingers to clip or wrap them. For tiny objects that cannot be enveloped in the hand’s palm, the MCP roll joints of two fingers move toward each other in opposite directions, allowing the fingertips to precisely aim at the object. When grasping longer objects like ice-cream sticks, a three-finger method ensures stability. For heavier and larger objects, an enveloping grasp can be used, where the fingers and palm enclose the entire object to ensure a stable hold. This flexibility in grasping strategies highlights the MuxHand's adaptability to a wide range of object sizes, shapes, and weights.

\subsection{Dexterous Manipulation Experiments}
\begin{figure*}
	\centering
  \includegraphics[width=0.98\textwidth]{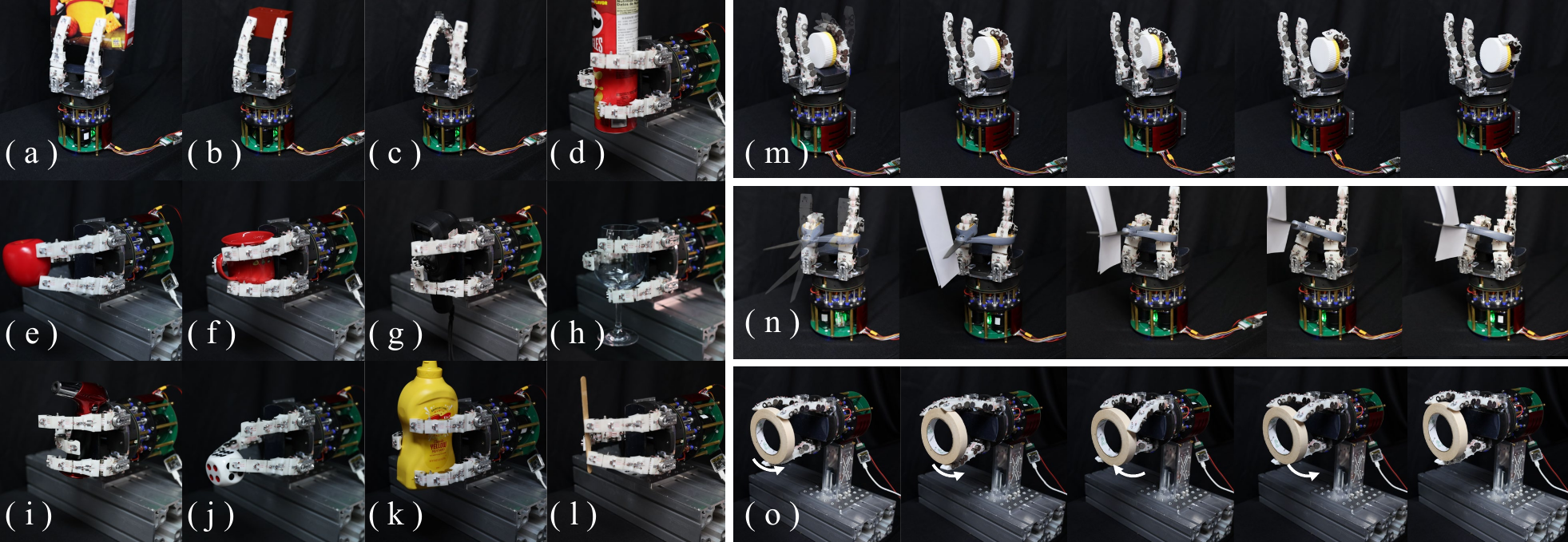}
	\caption{Grasping and dexterous manipulation experiments. (a)–(l) are the objects selected for grasping experiments : a cracker box, foam brick, USB hub, can of chips, apple, metal mug, thermometer, plastic wine glass, lighter, dice, mustard container, and ice-cream stick. (m) The hand rolls a disc. (n) The hand uses a pair of scissors to cut a sheet of paper. (o) The hand rotates a ring to a specific angle. } \label{fig:9}
  \vspace{-0.5cm}
\end{figure*}

Beyond grasping, dexterous manipulation is crucial for dexterous hand. We conduct several experiments to validate the MuxHand's dexterous manipulation capabilities. As shown in Fig. 7(m)-(o), the objects used for these experiments include a disc, a pair of scissors, and a ring.

In Fig. 7(m), the MuxHand uses its fingertip to press on a disc, rolling it back and forth within the palm. This demonstrates the hand's ability to finely adjust an object while maintaining continuous contact and control.
In the second experiment, shown in Fig. 7(n), the MuxHand uses a pair of scissors from the YCB object set to cut a sheet of paper. Mimicking human scissor usage, the scissors are mounted on the PIP joint. The MuxHand first grasps the scissors, then rotates the MCP roll joint to open and close them.
The final experiment, shown in Fig. 7(o), involves the MuxHand grasping a ring with its fingertips and rotating it to a specific angle. This demonstrates precise rotational manipulation.

\subsection{Fingertip load Experiments}
\begin{figure}
	\centering
  \includegraphics[width=0.48\textwidth]{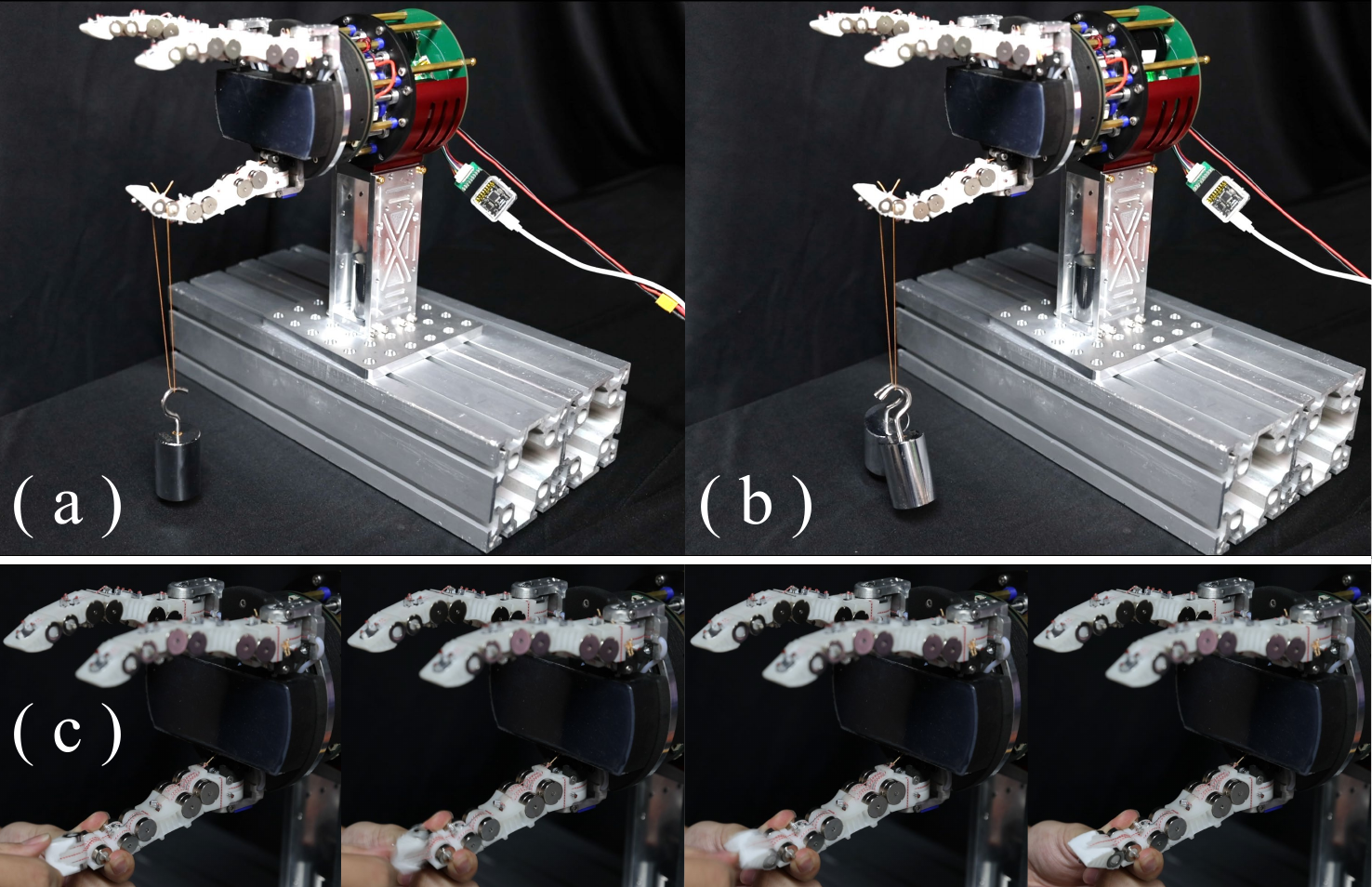}
	\caption{Fingertip load experiments. (a) Fingertip load of 500 g. (b) Fingertip load of 1.0 kg. (c) External forces are applied to the finger joints, and after removing the forces, the joints automatically reset to their original positions.} \label{fig:10}
 \vspace{-0.5cm}

\end{figure}

Additionally, we perform load experiments to evaluate the fingertip load capacity of the MuxHand. In tests, as shown in Fig. 8(a) and (b), we incrementally add a 500 g standard weight block to the fingertip of a single finger. The results show that the fingertip load capacity reached 1.0 kg. In addition, we apply external forces to the finger joints to deliberately cause joint misalignment. After removing the external forces, the fingers automatically reset to their initial positions due to the attraction of the radially magnetized magnets, as shown in Fig. 8(c). This experiment highlights the MuxHand’s self-resetting ability, ensuring stability and robustness even after experiencing external disturbances.

\section{Conclusion}

In this paper, we presented a cable-driven robotic hand named MuxHand, which utilizes the TDMM mechanism that we proposed. The TDMM mechanism enables the MuxHand to control 9 DOF across its three fingers using only 4 motors, greatly reducing both the system’s volume and cost. A series of experiments were conducted to validate MuxHand’s performance. The results shown that, through the TDMM mechanism and decoupled magnetic joints, the MuxHand achieves robust grasping and dexterous manipulation, with a fingertip load capacity of up to 1.0 kg. The magnetic joints further enhance durability, allowing the fingers to withstand external forces and impacts without damage.
In future work, we plan to improve the stability and grasping speed of the MuxHand, further enhancing its grasping and dexterous manipulation capabilities.

\bibliographystyle{ieeetr}
\bibliography{ref}

\vfill
\end{CJK}
\end{document}